\documentclass[letterpaper, 10pt, conference]{ieeeconf}

\IEEEoverridecommandlockouts
\usepackage{amsmath,amssymb,graphicx}
\graphicspath{{image/}}
\usepackage{booktabs,multirow,threeparttable,adjustbox}
\usepackage{algorithm}
\usepackage{algpseudocode}
\usepackage{dblfloatfix}
\usepackage{indentfirst}
\usepackage{subcaption}

\usepackage[pdftex]{hyperref}
\hypersetup{pdfauthor={},pdftitle={},pdfsubject={},pdfkeywords={}}

\usepackage{xurl}
\newcommand{\code}[1]{\texttt{\nolinkurl{#1}}}

\title{\LARGE \bf
Semantic Color Naturalness Breaker: Preventing Illegitimate Colorization via Content-Aware Color Priors
}

\author{Yuki Nii$^{\star}$ \qquad Futa Waseda$^{\star}$ \qquad Ching-Chun Chang$^{\dagger}$ \qquad Isao Echizen$^{\star\dagger}$%
\thanks{This work was partially supported by JSPS KAKENHI Grants JP21H04907 and JP24H00732, by JST CREST Grants JPMJCR20D3 and JPMJCR2562 including AIP challenge program, by JST AIP Acceleration Grant JPMJCR24U3, and by JST K Program Grant JPMJKP24C2 Japan.}%
\thanks{$^{\star}$The University of Tokyo, Japan}%
\thanks{$^{\dagger}$National Institute of Informatics, Japan}%
}

\begin{document}

\maketitle
\thispagestyle{empty}
\pagestyle{empty}

\begin{abstract}
Automatic image colorization enables large-scale and low-cost reuse of grayscale media (e.g., manga panels and archival photographs), facilitating unauthorized reuse and redistribution. Once released online, grayscale content can be readily turned into unauthorized colorized derivatives using off-the-shelf models, creating a practical need for proactive, content-side protection at publication time.
Building on \emph{Uncolorable Examples (UE)}, which add imperceptible perturbations to released grayscale images to degrade unauthorized colorization, we propose \textbf{Semantic Color Naturalness Breaker (SCNB)}---a semantic-level UE framework that drives colorization outputs toward \emph{content-inconsistent} colors while preserving the visual fidelity of the released grayscale media.
We further introduce \textbf{Content-aware Color Distributional Distance (CaCDD)}, a ground-truth-free, content-aware measure of color plausibility derived from semantic color priors, used both as the optimization objective of SCNB and as an evaluation metric.
Experiments on ImageNet show that our method remains effective under small perturbation budgets and common post-processing, supporting practical deployment in real-world content-sharing pipelines.
\end{abstract}

\section{INTRODUCTION}
Recent advances in generative-model-based automatic image colorization can produce realistic, diverse color images from a single grayscale input~\cite{wu2022vividdiverseimagecolorization,kang2023ddcolorphotorealisticimagecolorization,saharia2022palette}.
While beneficial for film restoration and content production, they also enable unauthorized colorization of monochrome media (e.g., manga panels and archival photographs), thereby lowering the cost of copyright-infringing redistribution and historical-image manipulation.
This is no longer hypothetical: unauthorized colorization is commoditized on freelance marketplaces~\cite{fiverr_colorize_gig}, and a recent report described an arrest involving sales of pirated, AI-colorized copies of a classic film~\cite{coda_ai_colorized_piracy_2026}.

Despite the growing misuse of automatic colorization, existing countermeasures remain largely on \emph{reactive}, post-hoc responses (detection/forensics, takedown, policy).
However, reactive defenses are costly and late: they act after dissemination and require continuous monitoring and enforcement.

\begin{figure}[t]
 \centering
 \includegraphics[width=\columnwidth]{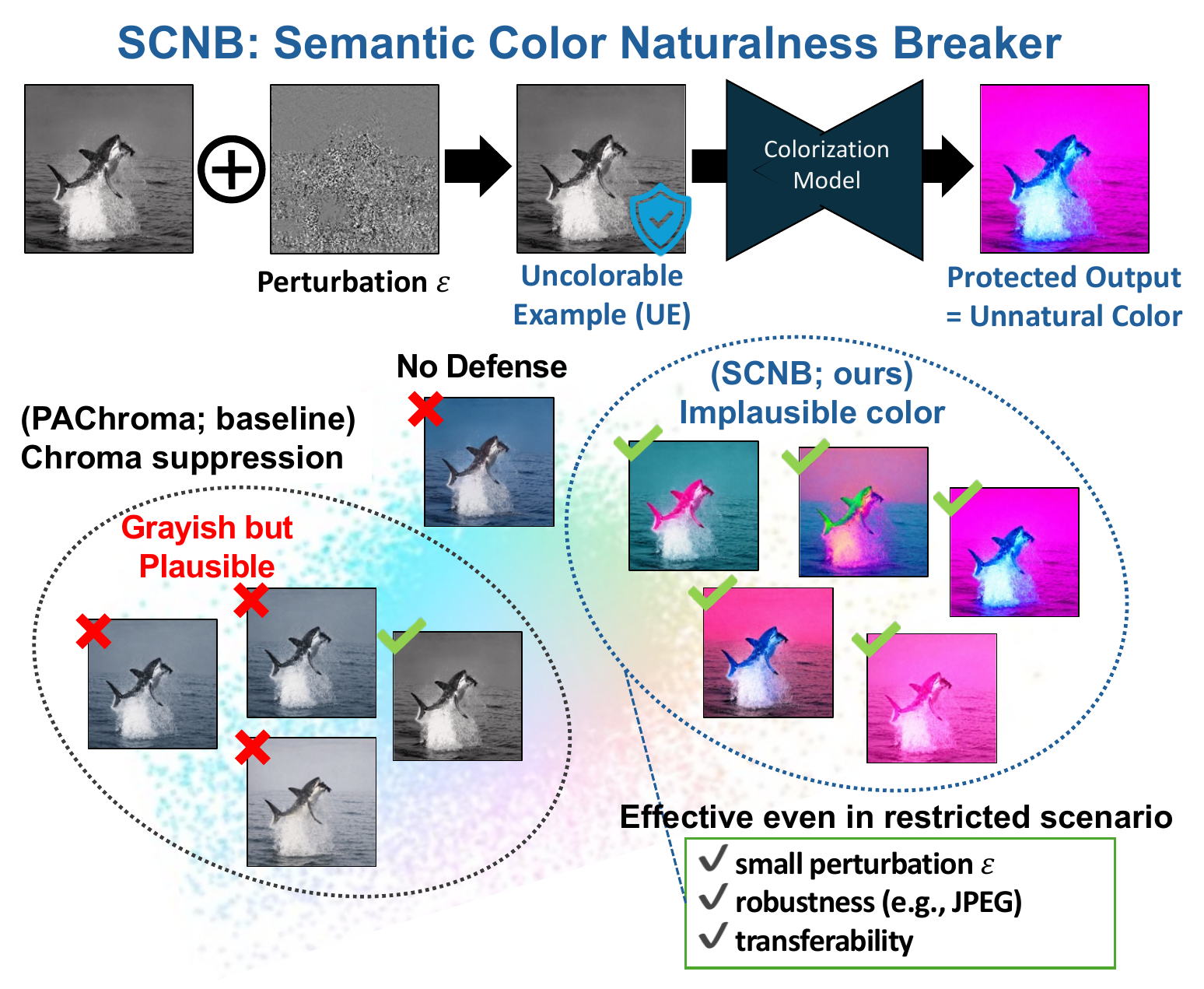}
\caption{\textbf{SCNB breaks semantic color plausibility.}
UE generated from PAChroma suppresses chroma magnitude but can leave \emph{gray-but-natural} outputs, whereas SCNB induces \emph{content-inconsistent} colors even under restricted settings.}
  \label{fig:teaser}
\end{figure}

We therefore pursue a \emph{proactive, content-side defense} at publication time: the provider releases grayscale media that preserves utility for viewing and legitimate use, while making unauthorized colorization \emph{semantically implausible}---i.e., the colorized outputs contain scene-inconsistent colors.
Prior work introduced the concept of \textbf{Uncolorable Examples (UE)}~\cite{nii2025uncolorable} as a content-side defense, and proposed a chroma-suppression method, \textbf{PAChroma}~\cite{nii2025uncolorable}, which aimed to enforce grayscale outputs.
However, we identify a fundamental limitation: under realistic constraints, including limited perturbation budgets and common post-processing, PAChroma often yields intermediate near-grayscale outputs that remain \emph{gray-but-natural} (Fig.~\ref{fig:teaser}).
Crucially, such outputs can retain \emph{faint yet content-consistent} colors that remain visually plausible, and thus insufficient for practical defense.

\begin{figure*}[t]
\centering
\includegraphics[width=\linewidth]{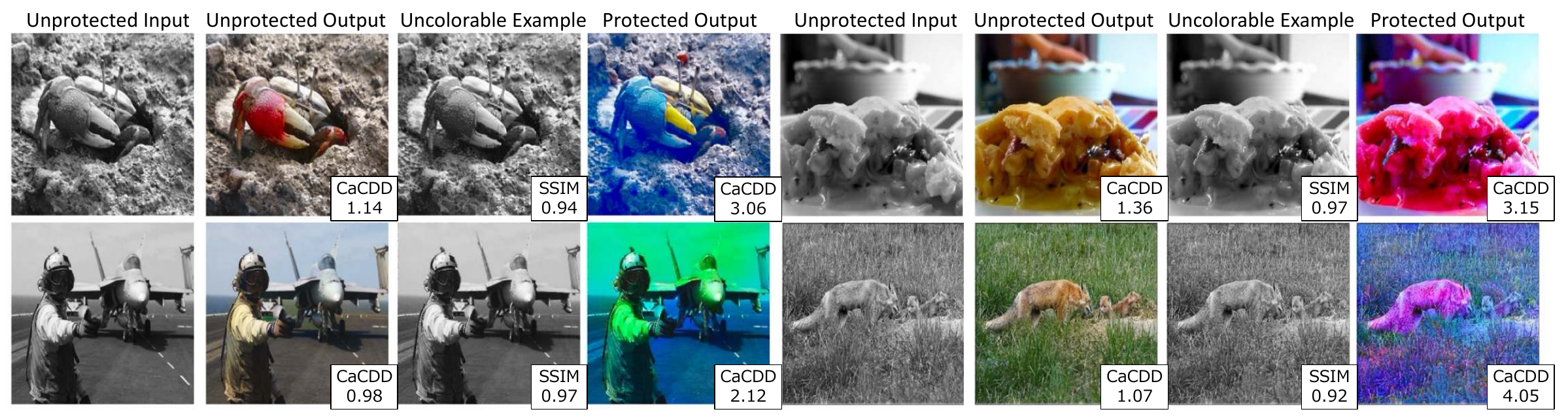}
\caption{\textbf{Uncolorable Example generated from SCNB inducing implausible color.} Each image is shown with its SSIM between the inputs and CaCDD score.}
\label{fig:qualitative}
\end{figure*}

To overcome this fundamental limitation, in this work, we introduce a semantic-level defense, which targets output plausibility by driving colorization toward \emph{content-inconsistent} colors. Our contributions are summarized as follows:

\begin{itemize}
\item \textbf{Semantic-level content-side defense:} We propose \emph{Semantic Color Naturalness Breaker (SCNB)}, an Uncolorable Example (UE) framework that preserves the utility of released grayscale media while inducing \emph{content-inconsistent} color in unauthorized colorizations.
\item \textbf{Semantic color plausibility metric (CaCDD):} We introduce \emph{Content-aware Color Distributional Distance (CaCDD)}, which quantifies color plausibility using \emph{semantic priors} and serves both as the optimization objective of SCNB and as a standalone evaluation metric.
\item \textbf{Real-world suitability:} Experiments on ImageNet show that SCNB remains effective under small perturbation budgets and common post-processing operations, supporting practical deployment in real-world sharing pipelines.
\end{itemize}

\section{RELATED WORK}
\textbf{Automatic colorization.}
Automatic colorization has evolved from deterministic regression to multimodal, prior-driven generation.
With large-scale data and expressive models, recent systems produce diverse, high-fidelity colors via GANs, transformers,
and learned perceptual priors~\cite{kang2023ddcolorphotorealisticimagecolorization,DeOldify2021,kim2022bigcolorcolorizationusinggenerative}.
We target a broad range of modern colorizers, treating them as potentially untrusted tools for unauthorized use.

\textbf{Adversarial examples and proactive protection.}
Adversarial perturbations can reliably influence neural outputs~\cite{szegedy2013intriguing,goodfellow2014explaining,madry2018pgd}, and have been extended beyond classification to generative models, including transformation-consistent formulations that improve robustness and transferability~\cite{shen2021structure}.
Recently, proactive content protection has been explored to disrupt downstream misuse, including defenses against generative imitation or editing (e.g., artist/style protection)~\cite{shan2023glaze,shan2024nightshade,salman2023photoguard}.
Our work targets a distinct misuse channel: unauthorized automatic colorization of grayscale media. We study \emph{Uncolorable Examples (UE)}~\cite{nii2025uncolorable}, and propose a new objective that breaks \emph{semantic color plausibility. }

\textbf{Color plausibility metrics.}
Existing metrics largely assess \emph{image-level} quality rather than \emph{content-semantics--conditioned} color plausibility.
PSNR/SSIM require a single ground-truth and quantify pixel/structure fidelity.
LPIPS and FID evaluate perceptual similarity or feature-space realism, but they do not explicitly test whether the \emph{assigned color is appropriate for the semantic content}.
Chroma-magnitude measures (e.g., colorfulness~\cite{colorfulness}) capture saturation strength, yet ignore semantic correctness.
We address this gap with CaCDD, a differentiable, ground-truth-free metric that measures color plausibility by comparing predicted colors against semantic priors, enabling both principled evaluation and direct optimization for SCNB.

\section{UNCOLORABLE EXAMPLES}
\textbf{Threat model and design goals.}
\label{subsec:threat_model}
We consider a content provider (defender) who publishes a grayscale image $x_l$ and wants to discourage unauthorized colorization by third-party colorizers.
The defender releases a protected image $x^{\text{adv}} = x_l + \delta$, where $\delta$ is an additive, imperceptible perturbation constrained by a small budget.
This protection preserves the utility of the grayscale content while causing colorizers to produce outputs unsuitable for reuse or republishing.
The attacker may choose any colorization model $G(\cdot)$ and may apply lightweight post-processing commonly introduced during online sharing.

Following prior work~\cite{nii2025uncolorable}, UE should meet four requirements for practical protection:
\textbf{(1) Effectiveness} (degrade colorization outputs),
\textbf{(2) Imperceptibility} (preserve original appearance),
\textbf{(3) Robustness} (remain effective after post-processing), and
\textbf{(4) Transferability} (work on unseen colorizers when crafted on a surrogate $G_s$).

\section{METHOD}
\subsection{Motivation: From Chroma Suppression to Semantic Inconsistency}
PAChroma~\cite{nii2025uncolorable} is a chroma-restrictive UE baseline that suppresses chroma generation.
\label{subsec:pachroma_limitation}
In the ideal case of PAChroma, sufficiently strong suppression would collapse chroma.
In practice, however, proactive defenses often operate under realistic constraints---small perturbation budgets for imperceptibility, distribution shifts due to platform post-processing, and imperfect transfer across diverse colorization models.
Under these conditions, chroma suppression becomes only \emph{moderately} effective: instead of fully neutralizing chroma, outputs often fall into an intermediate regime between grayscale and vivid colorization (Fig.~\ref{fig:teaser}).
In practice, they retain faint but \emph{content-consistent} color (e.g., bluish skies or greenish vegetation), yielding \emph{near-grayscale} yet usable results for illegitimate coloring.
We call this common failure mode \emph{gray-but-natural}.
Existing approaches implicitly treat reduced chroma magnitude as success, but this overlooks a significant limitation: suppressing chroma does not reliably break \emph{semantic color plausibility}.

This motivates \textbf{SCNB}.
Rather than minimizing chroma magnitude, SCNB explicitly drives the output color to become \emph{inconsistent with content-semantics--conditioned color priors}, producing colorizations that are difficult for downstream reuse.
Fig.~\ref{fig:qualitative} visualizes the resulting content-inconsistent colorizations produced by SCNB.

\subsection{SCNB: Semantic Color Naturalness Breaker}
\label{sec:SCNB}
SCNB crafts protected grayscale inputs that make unauthorized colorizers produce content-inconsistent colors.
Given a grayscale image $x_l$ and a target colorizer $G$, we optimize a bounded perturbation $\delta$ to maximize a content-aware color score
$\mathrm{CaCDD}(\cdot)$ (Sec.~\ref{sec:CaCDD}):

\begin{equation}
\delta^{\star}
=
\arg\max_{\|\delta\|_\infty \le \epsilon}
\ \mathrm{CaCDD}\!\big(G(x_l + \delta)\big).
\label{eq:SCNB_opt}
\end{equation}

\begin{figure}[t]
  \centering
 \includegraphics[width=\columnwidth]{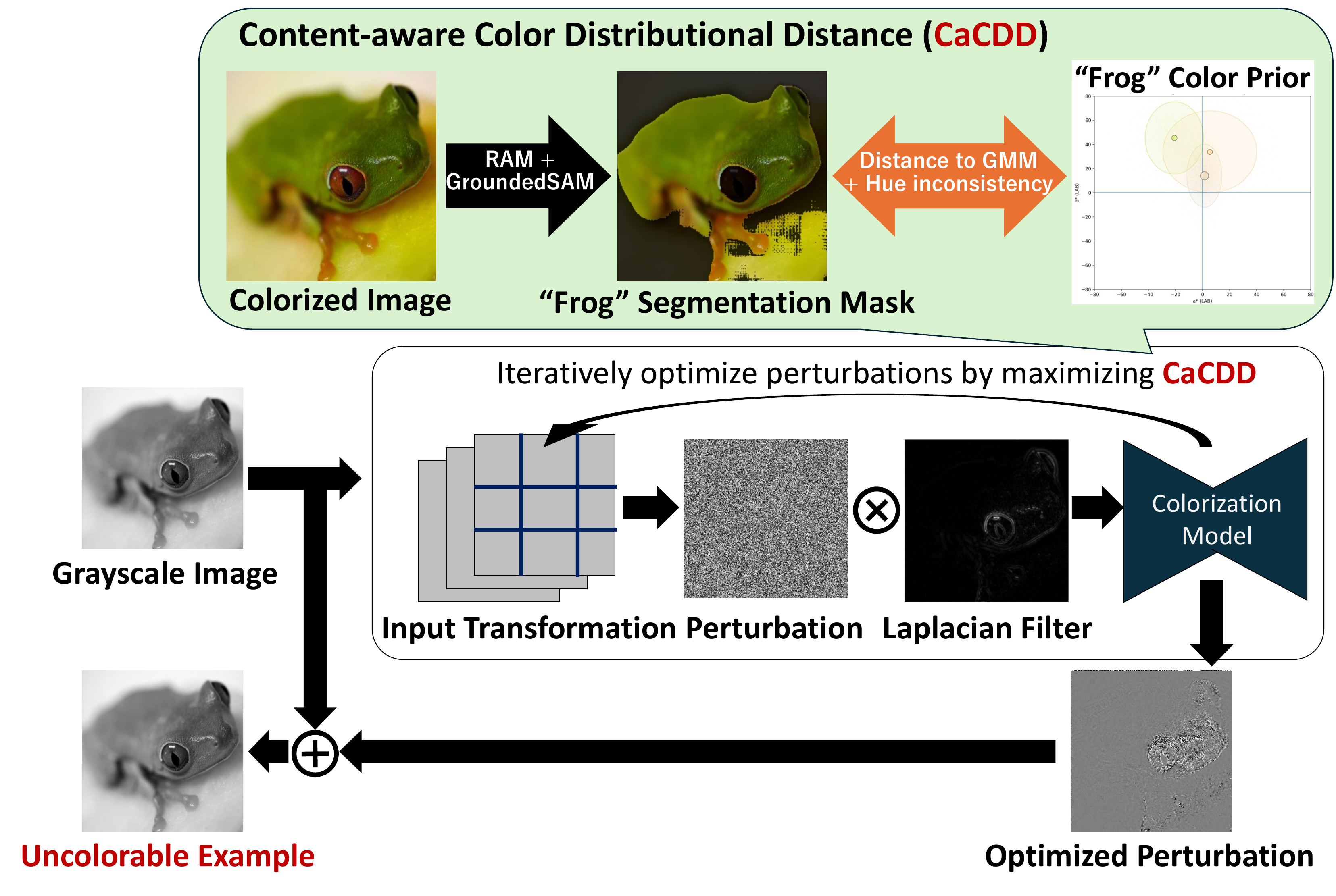}
 \caption{\textbf{SCNB pipeline.}  SCNB maximizes CaCDD to induce \emph{content-inconsistent} color via adversarial examples utilized by input transformation and a Laplacian mask.}
  \label{fig:pipeline}
\end{figure}

In practice, SCNB follows the same optimization backbone as PAChroma: a perception-aware masked MI-FGSM update using
(i) a Laplacian-based imperceptibility mask and
(ii) input transformation (random shift/scale/resize-crop and block-wise variants) to improve robustness and transferability.
Alg.~\ref{alg:SCNB} and Fig.~\ref{fig:pipeline}
summarizes the procedure.

\begin{algorithm}[t]
\small
\caption{Semantic Color Naturalness Breaker (SCNB)}
\label{alg:SCNB}
\textbf{Input:} Colorization model $G(\cdot)$; CaCDD score $\mathrm{CaCDD}(\cdot)$ (Sec.~\ref{sec:CaCDD});
grayscale luminance $x_l$; perturbation budget $\epsilon$; iterations $T$; momentum decay $\mu$;
block split $s$; number of transformations $N$; perception-aware Laplacian mask $M$ \\
\textbf{Output:} Protected image $x^{\text{adv}}_T$
\begin{algorithmic}[1]
\State Set step size $\alpha \leftarrow \epsilon/10$; initialize $g_0 \leftarrow 0$, $x^{\text{adv}}_0 \leftarrow x_l$
\For{$t=0$ to $T-1$}
    \State Sample input transformations $\{\mathcal{T}_i\}_{i=1}^N$ (with block split $s$)
    \For{$i=1$ to $N$}
        \State $x^{\text{tran}}_i \leftarrow \mathcal{T}_i(x^{\text{adv}}_t)$
        \State $y_i \leftarrow G(x^{\text{tran}}_i)$
        \State $g^{(i)} \leftarrow \nabla_{x^{\text{tran}}_i}\ \mathrm{CaCDD}(y_i)$ \Comment{maximize CaCDD to induce color implausibility}
    \EndFor
    \State $\bar{g}_{t+1} \leftarrow \frac{1}{N}\sum_{i=1}^N g^{(i)}$
    \State $g_{t+1} \leftarrow \mu\, g_t + \bar{g}_{t+1}/\|\bar{g}_{t+1}\|_1$
    \State $\Delta \leftarrow M \cdot \alpha \cdot \mathrm{sign}(g_{t+1})$
    \State $x^{\text{adv}}_{t+1} \leftarrow \mathrm{Clip}_{[x_l-\epsilon,\ x_l+\epsilon]}\!\left(x^{\text{adv}}_t + \Delta\right)$
\EndFor
\State \Return $x^{\text{adv}}_T$
\end{algorithmic}
\end{algorithm}

\begin{figure}[t]
\centering
\includegraphics[width=\columnwidth]{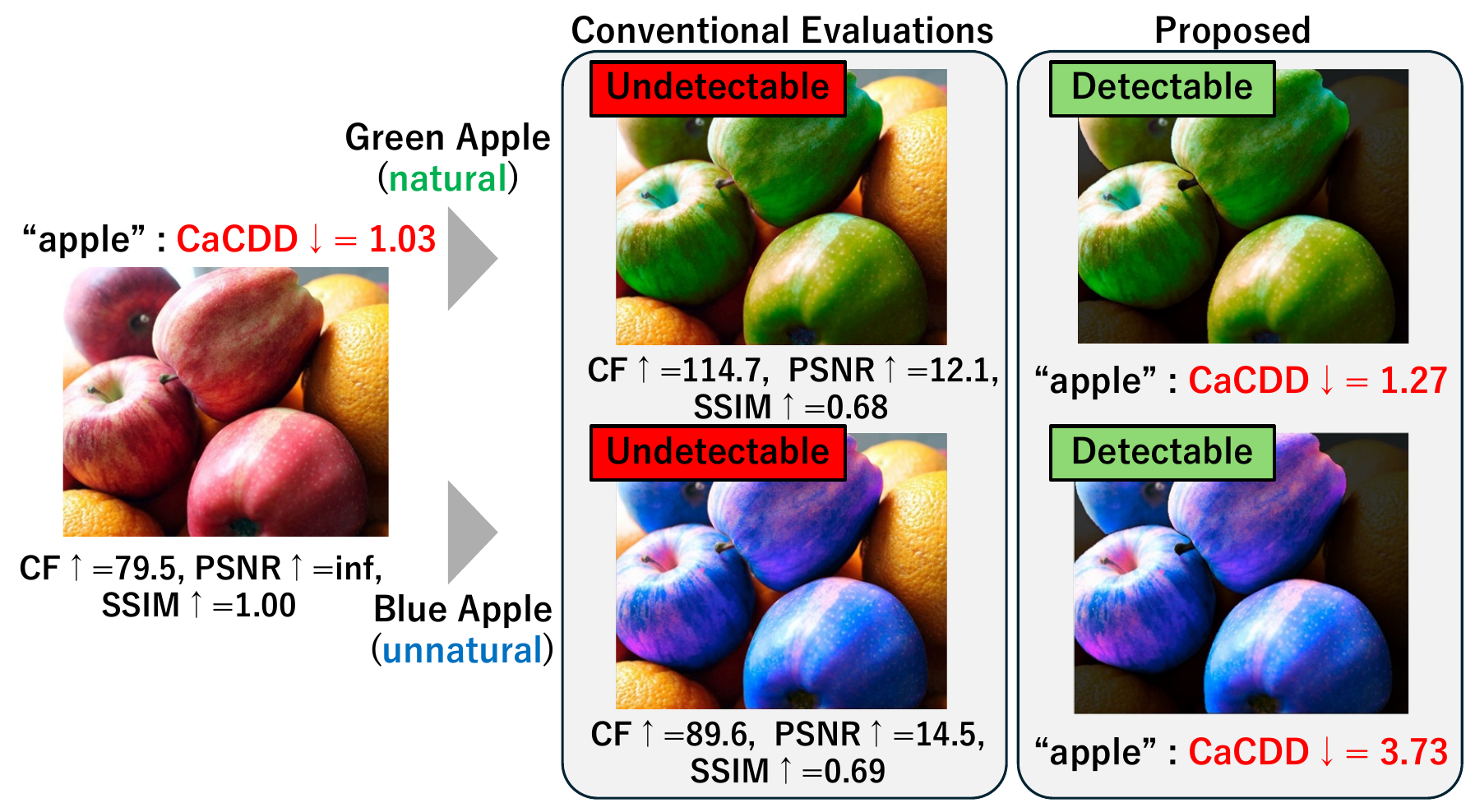}
\caption{\textbf{CaCDD captures content-aware color plausibility.} Distinguishing natural from semantically implausible colorizations beyond CF/PSNR/SSIM.}
\label{fig:metric_fail_apple}
\end{figure}

\subsection{CaCDD}
\label{sec:CaCDD}

\textbf{Intuition.}
Standard chroma and perceptual metrics such as colorfulness (CF)~\cite{colorfulness}, PSNR, and SSIM evaluate image-level distortion or saturation magnitude, but do not account for whether the predicted color is semantically appropriate for the depicted content (e.g., a blue apple would score similarly to a natural image if chroma magnitude and structure are preserved).
Content-aware Color Distributional Distance (CaCDD) addresses this gap as a ground-truth-free score that measures how far the colorization deviates from a label-conditioned semantic color prior (Fig.~\ref{fig:metric_fail_apple}).

\begin{figure}[t]
\centering
\includegraphics[width=0.75\linewidth]{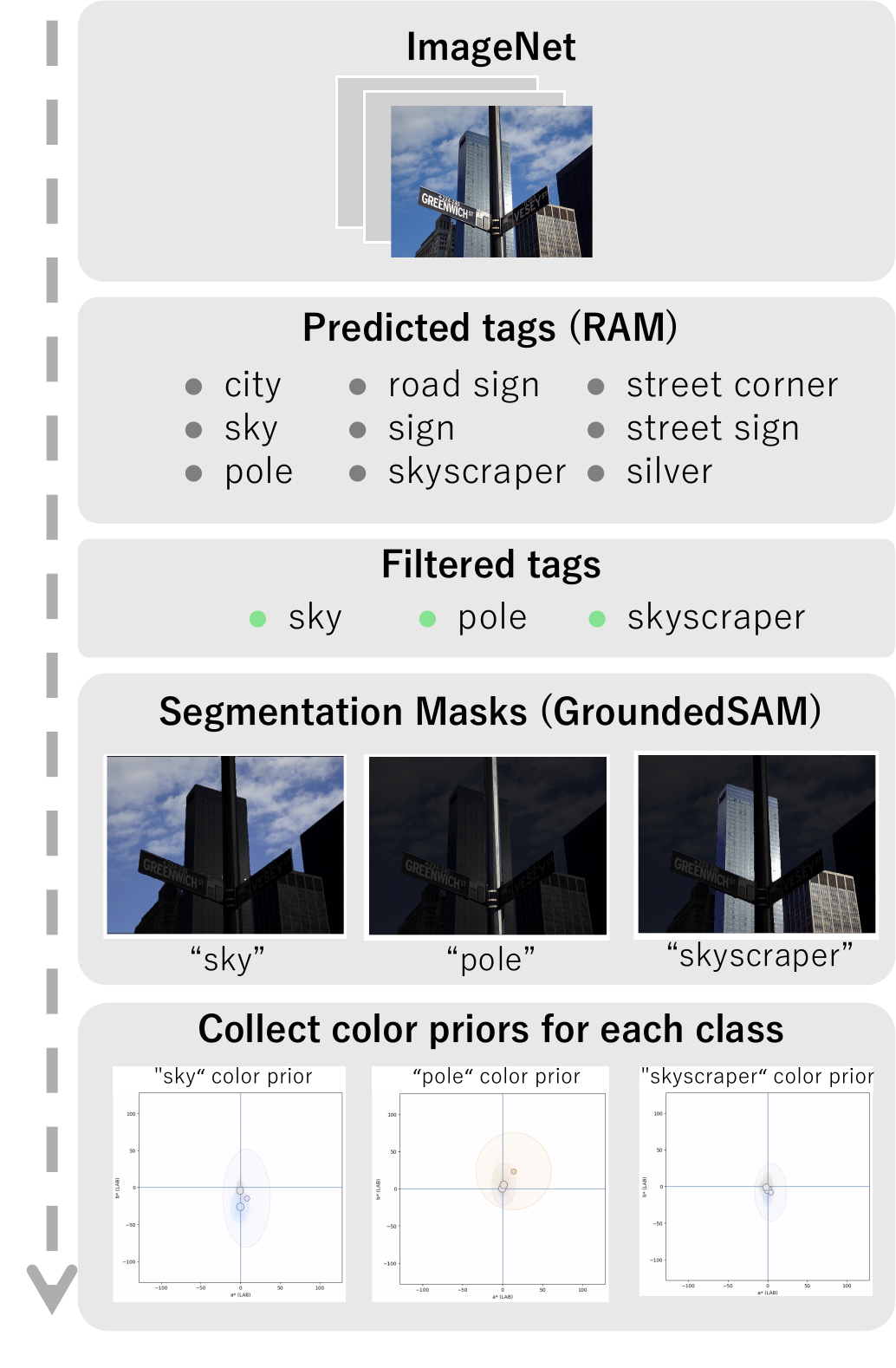}
\caption{\textbf{Semantic prior construction.} Open-vocabulary tags and masks are used to collect ab channels of CIELAB per class and fit class-conditional GMM priors.}
\label{fig:supp_prior_pipeline}
\end{figure}

\textbf{Semantic color priors.}
For each semantic label $c\in\mathcal{C}$, we construct a color prior by fitting
a $K$-component GMM $p_{\text{real}}^{(c)}(z)$ over chroma samples
$z=(a,b)\in\mathbb{R}^2$ in CIELAB space ($K{=}3$ unless noted).
To build these priors at scale without manual annotation, we employ an
open-vocabulary tagging-segmentation pipeline (Fig.~\ref{fig:supp_prior_pipeline}):
for each reference image, RAM~\cite{zhang2023ram} predicts a ranked list of
semantic tags as candidate labels, and Grounded-SAM~\cite{ren2024groundedsam}
localizes each tag into a pixel-accurate binary mask via text-grounded
detection and segmentation.
We discard low-quality masks using geometric and confidence-based filters
(e.g., minimum area ratio and grounding score thresholds), and collect
$(a,b)$ color samples from the remaining reliable regions.
The aggregated samples are then used to fit a per-label GMM.
For evaluation, we compute semantic masks once per input and reuse them,
fixing the label assignment $c(p)$ for each pixel $p$.

\begin{figure}[t]
\centering
\includegraphics[width=0.9\columnwidth]{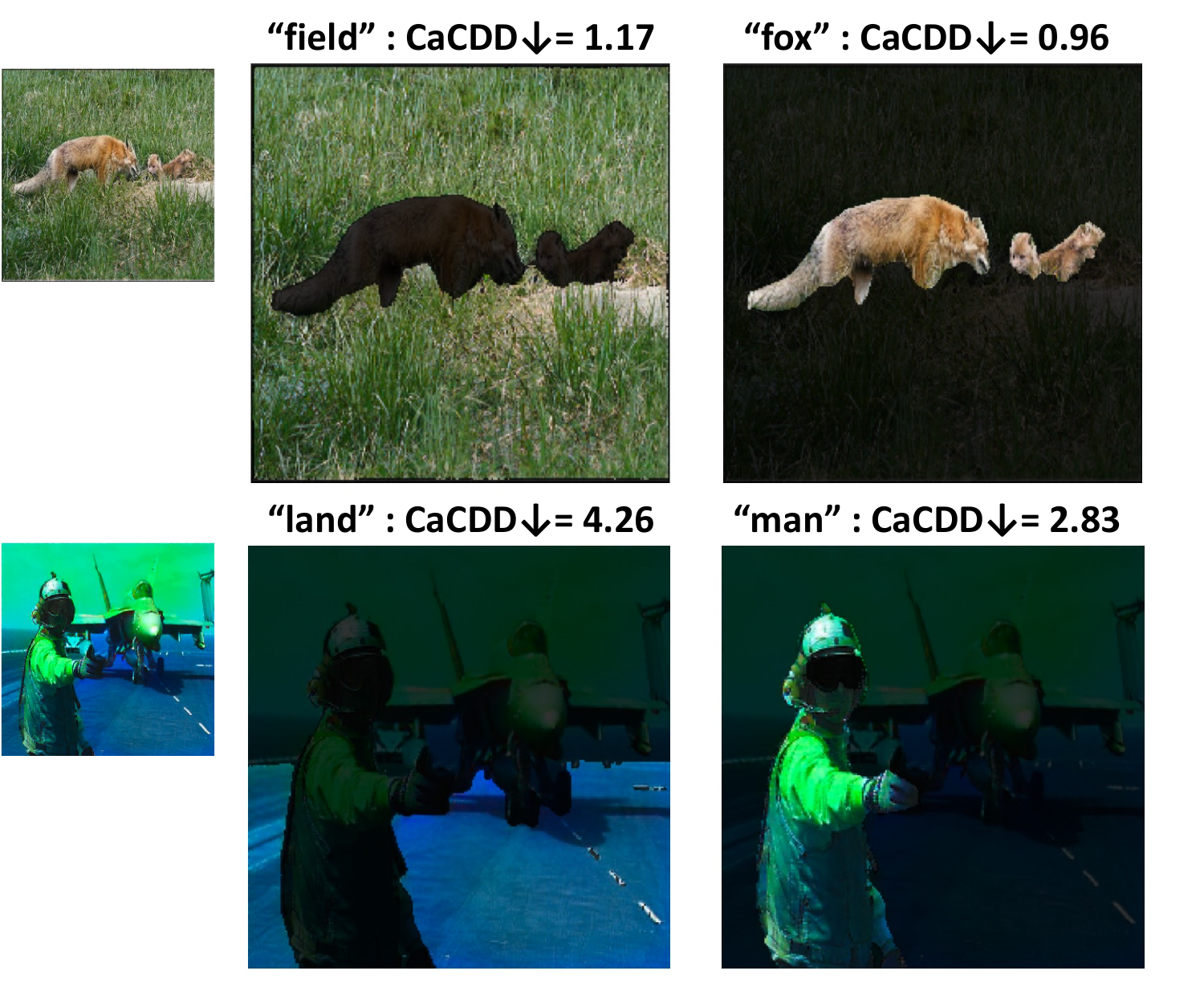}
\caption{\textbf{Region-wise CaCDD visualization.} Low CaCDD indicates semantically plausible colors, whereas high CaCDD indicates content-inconsistent colors.}
\label{fig:supp_region_natural}
\end{figure}

\textbf{Per-pixel CaCDD.}
Given a colorization $y$, let $z_p=(a_p,b_p)$ be the chroma at pixel $p$ and $c(p)$ its semantic label.
CaCDD scores each pixel by combining two complementary terms:
(i) \emph{OoD chroma distance}, which measures how far $z_p$ lies from the label-conditioned GMM in Mahalanobis distance, capturing cases where the predicted color falls outside the plausible color distribution for that semantic class; and (ii) \emph{hue-direction inconsistency}, which penalizes pixels whose chroma direction in the $ab$ plane diverges from the nearest GMM mean direction, capturing cases where the chroma magnitude may be moderate but the hue itself is wrong (e.g., reddish vegetation instead of green).
Let $\{\pi_{c,k},\mu_{c,k},\Sigma_{c,k}\}_{k=1}^{K}$ be the GMM parameters, $\varepsilon_d$ a small constant for numerical stability, $\tau$ a chroma threshold below which hue is unreliable, and $\lambda_{\text{hue}}$ the weight balancing the two terms.
Fig.~\ref{fig:supp_region_natural} reports CaCDD scores per semantic class for each image: a low CaCDD score indicates semantically plausible colors.

\textit{(i) OoD chroma distance.}
We first define a responsibility-weighted squared Mahalanobis distance
\begin{equation}
D(p)=\sum_{k=1}^{K} r_{p,k}\, d_{p,k}^2,
\end{equation}
where $r_{p,k}=p(k\mid z_p,c(p))$ is the posterior responsibility under the label-conditioned GMM:
\begin{equation}
r_{p,k}=\frac{s_{p,k}}{\sum_{j=1}^{K}s_{p,j}},\quad
s_{p,k}=\pi_{c(p),k}\,\mathcal{N}\!\left(z_p;\mu_{c(p),k},\Sigma_{c(p),k}\right),
\end{equation}
\begin{equation}
d_{p,k}^2=(z_p-\mu_{c(p),k})^{\top}\Sigma_{c(p),k}^{-1}(z_p-\mu_{c(p),k}).
\end{equation}
We apply a log transform to stabilize the term, $\log(1+D(p))$.

\textit{(ii) Hue-direction inconsistency.}
We define a hue inconsistency score
\begin{equation}
H(p)= w_p\, \min\!\Big(1,\frac{\theta_p}{\pi/2}\Big),
\end{equation}
where $\theta_p$ is the angle between the pixel color direction and the nearest GMM mean direction, and
$w_p=\min(1,\|z_p\|/\tau)$ down-weights near-grayscale pixels.
Let
\begin{equation}
\begin{aligned}
m_p &= \max_{k}\hat{z}_p^{\top}\hat{\mu}_{c(p),k},\\
\hat{z}_p &= \frac{z_p}{\|z_p\|+\varepsilon_d},\\
\hat{\mu}_{c(p),k} &= \frac{\mu_{c(p),k}}{\|\mu_{c(p),k}\|+\varepsilon_d},
\end{aligned}
\end{equation}
and set $\theta_p=\arccos(\operatorname{clip}(m_p,-1,1))$.

\textbf{Final CaCDD.}
The two terms are combined into a single per-image score by averaging over all semantically labeled pixels:
\begin{equation}
\mathrm{CaCDD}(y)=\frac{1}{|\Omega(y)|}\sum_{p\in\Omega(y)}\Big(\log(1+D(p))+\lambda_{\text{hue}}\,H(p)\Big),
\end{equation}
where $\Omega(y)$ denotes pixels within accepted semantic masks.

\begin{table*}[t]
\centering
\caption{\textbf{ImageNet transferability and robustness (BigColor as source).} ``Source'' crafts the perturbation and ``Attack'' is used for evaluation (white-box: Source$=$Attack). $\epsilon{=}32/255$. 
}
\label{tab:imagenet_transfer_scnb_pachroma_bigcolor_source}
\footnotesize
\setlength{\tabcolsep}{1.6pt}
\renewcommand{\arraystretch}{1.05}
\resizebox{\linewidth}{!}{%
\begin{tabular}{lll ccccc ccccc cc}
\toprule
\multirow{2}{*}{Method} & \multirow{2}{*}{Source model} & \multirow{2}{*}{Attack model} &
\multicolumn{5}{c}{CF (desaturation indicator)} &
\multicolumn{5}{c}{CaCDD $\uparrow$} &
\multicolumn{2}{c}{Output $\downarrow$} \\
\cmidrule(lr){4-8}\cmidrule(lr){9-13}\cmidrule(lr){14-15}
& & &
Unprot. & Prot. & JPEG75 & JPEG50 & RRC &
Unprot. & Prot. & JPEG75 & JPEG50 & RRC &
PSNR & SSIM \\
\midrule

\multirow{3}{*}{PAChroma}
& \multirow{3}{*}{BigColor}
& DeOldify & 36.12 & \textbf{21.33 (-41\%)} & 20.89 & 21.36 & 20.42 & 1.41 & \textbf{1.50 (+6\%)}  & 1.57 & 1.58 & 1.43 & 23.44 & 0.80 \\
& & BigColor & 31.44 & \textbf{4.13 (-87\%)}  & 5.42  & 7.16  & 4.96  & 1.26 & \textbf{1.05 (-17\%)} & 1.07 & 1.06 & 1.03 & 21.62 & 0.74 \\
& & DDColor  & 37.98 & \textbf{25.99 (-32\%)} & 24.30 & 24.10 & 26.10 & 1.25 & \textbf{1.36 (+9\%)}  & 1.28 & 1.28 & 1.38 & 21.62 & 0.74 \\

\midrule

\multirow{3}{*}{SCNB}
& \multirow{3}{*}{BigColor}
& DeOldify & 36.12 & \textbf{28.82 (-20\%)} & 26.91 & 25.82 & 28.05 & 1.41 & \textbf{1.62 (+16\%)} & 1.64 & 1.60 & 1.56 & 23.71 & 0.81 \\
& & BigColor & 31.44 & \textbf{76.67 (+144\%)} & 67.92 & 58.42 & 68.99 & 1.26 & \textbf{3.05 (+142\%)} & 2.91 & 2.68 & 2.89 & 14.27 & 0.57 \\
& & DDColor  & 37.98 & \textbf{32.82 (-14\%)} & 29.41 & 28.02 & 33.62 & 1.25 & \textbf{1.50 (+20\%)} & 1.43 & 1.39 & 1.50 & 21.47 & 0.76 \\

\bottomrule
\end{tabular}%
}
\end{table*}

\section{EXPERIMENTS}
We evaluate against three representative colorizers spanning different architectures and color generation strategies: DeOldify~\cite{DeOldify2021} (CNN based), BigColor~\cite{kim2022bigcolorcolorizationusinggenerative} (GAN-based with BigGAN priors), and DDColor~\cite{kang2023ddcolorphotorealisticimagecolorization} (dual-decoder transformer with multi-scale color decoding).
We build semantic color priors from the ImageNet training set~\cite{deng2009imagenet} by extracting RAM tags and Grounded-SAM masks, collecting per-region $(a,b)$ samples in CIELAB color space, and fitting per-label GMMs across 3{,}032 semantic labels.
Evaluation uses 40 ImageNet validation images with two random seeds.

We perturb the grayscale luminance under an $\ell_\infty$ budget $\epsilon \in \{32,16,8,4\}/255$.
SCNB uses $T{=}100$ iterations with step size $\alpha{=}\epsilon/10$, momentum decay $\mu{=}1.0$, and $N{=}20$ sampled transformations per iteration with block split $s{=}3$ (Alg.~\ref{alg:SCNB}).
We report colorfulness (CF)~\cite{colorfulness} and CaCDD on colorizer outputs to assess effectiveness.
Imperceptibility is measured using SSIM between the original and protected grayscale.
Robustness is evaluated under JPEG compression (quality 75 and 50) and random resized cropping (RRC), simulating common degradation encountered during online sharing.

All experiments were conducted on a single NVIDIA RTX 6000 GPU.
For the reported setting, processing 40 images required approximately 6 hours in total
(about 10 minutes per image on average).
Since protection is applied once at publication time and does not affect the viewing experience, this one-time cost is acceptable for practical deployment.

\subsection{Results and Observations}
\begin{figure}[t]
\centering
\includegraphics[width=\linewidth]{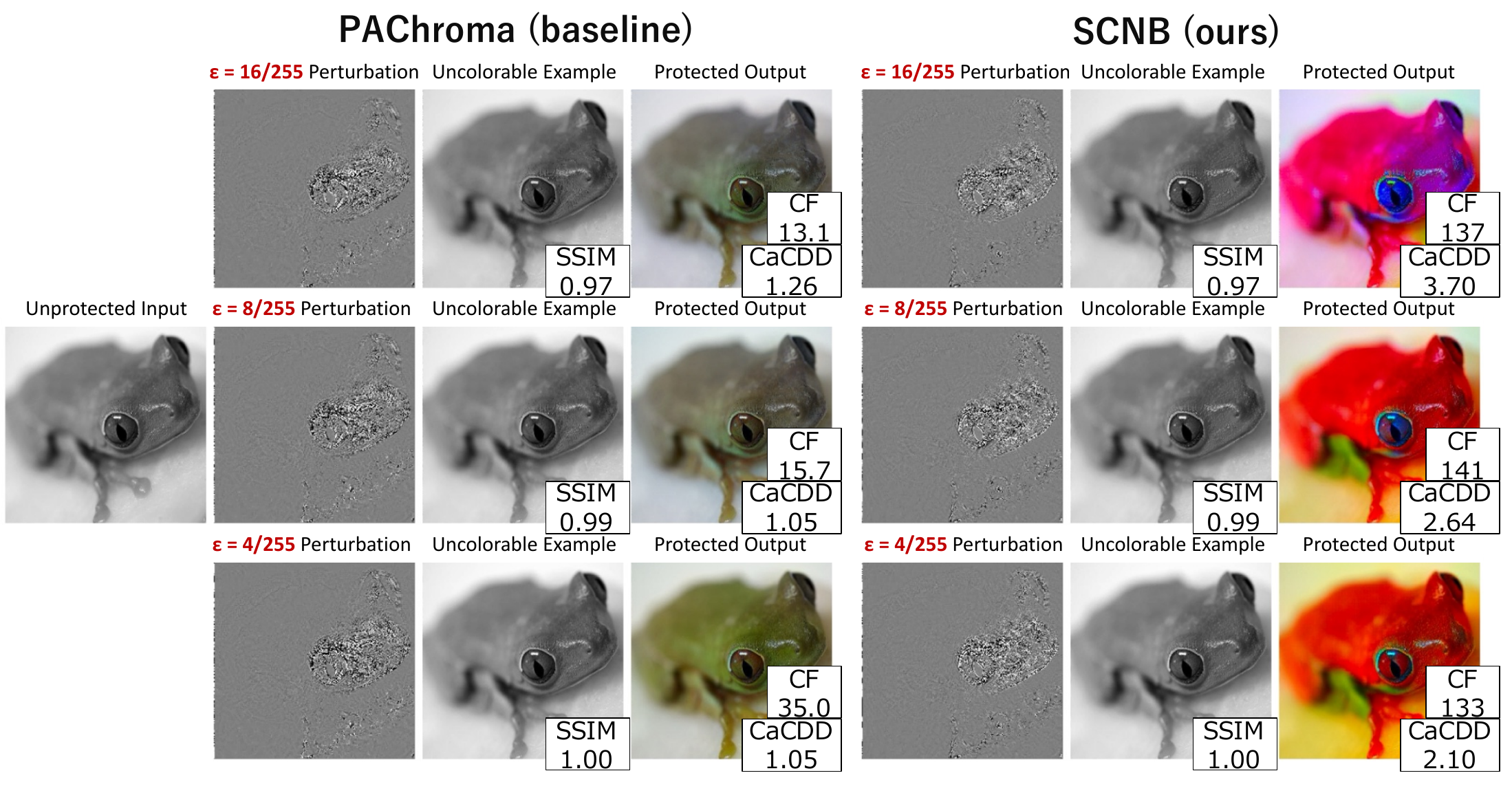}
\caption{\textbf{Effectiveness under small perturbation budgets.} As $\epsilon$ decreases, SCNB still induces content-inconsistent colors, whereas PAChroma tends to produce desaturated yet semantically plausible outputs. (Best viewed in zoom)}
\label{fig:epsilon_effect}
\end{figure}

\textbf{Effectiveness.}
Figure~\ref{fig:epsilon_effect} illustrates the qualitative difference between PAChroma and SCNB.
PAChroma drives the outputs toward near-grayscale; however, the residual faint color can remain \emph{content-consistent} (\emph{gray-but-natural}), so the colorization may still appear plausible despite desaturation (Sec.~\ref{subsec:pachroma_limitation}).
In contrast, SCNB induces \emph{visibly content-inconsistent} colors indicating a semantic break rather than simple chroma suppression.
Table~\ref{tab:imagenet_transfer_scnb_pachroma_bigcolor_source} corroborates this trend quantitatively.
While PAChroma reliably reduces CF, CaCDD often \emph{decreases} or changes only slightly, consistent with the \emph{gray-but-natural} failure mode.
SCNB, however, substantially increases CaCDD in the white-box setting across all three colorizers, e.g.,
BigColor $1.26\!\rightarrow\!3.05$ (+142\%),
showing that the outputs deviate from content-conditioned color priors.

\begin{figure}[t]
\centering
\includegraphics[width=\linewidth]{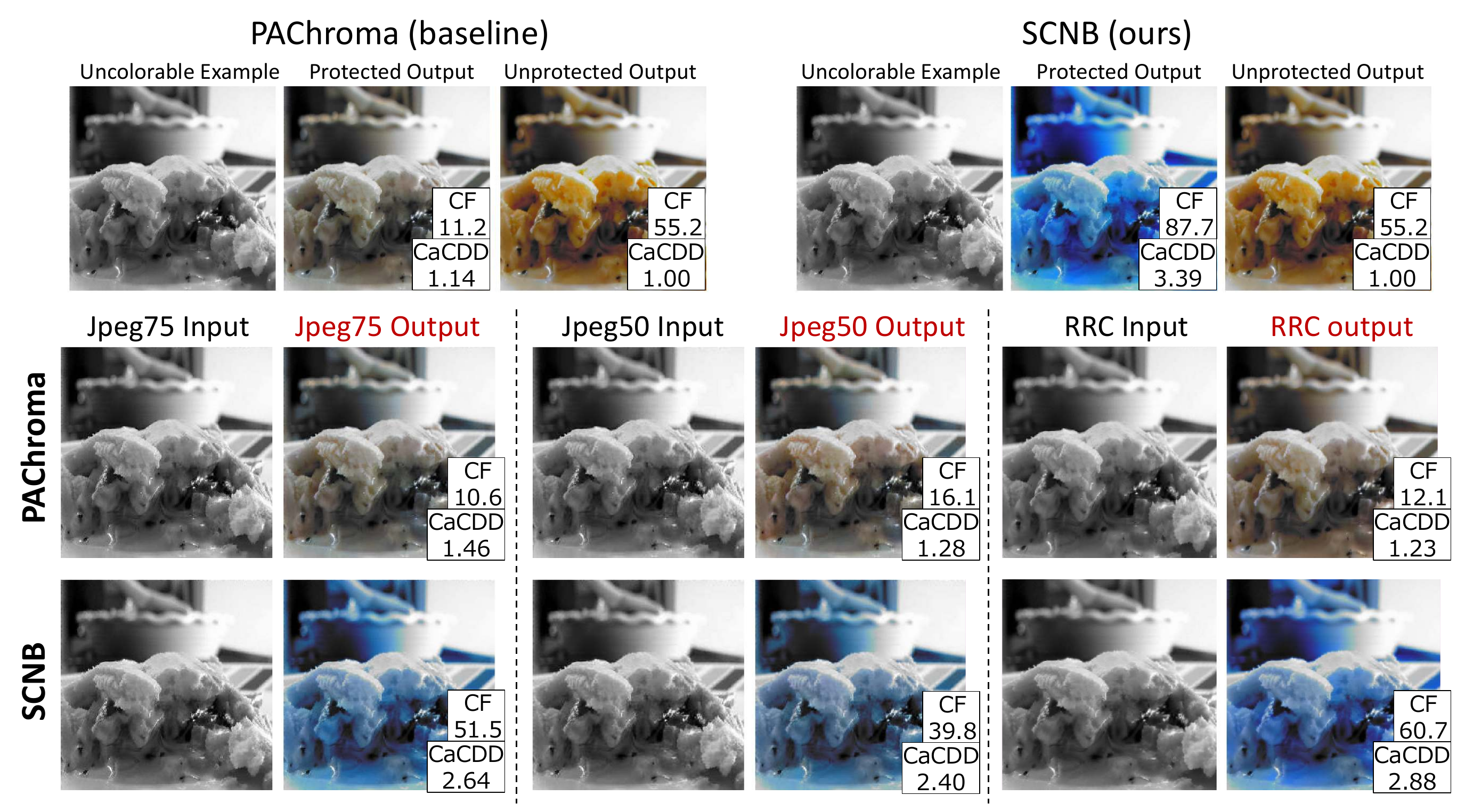}
\caption{\textbf{Robustness to common post-processing.} SCNB remains content-inconsistent after JPEG (Q=75/50) and random resized crop (RRC), whereas PAChroma tends to yield desaturated but still plausible colors. (Best viewed in zoom)}
\label{fig:robustness}
\end{figure}

\textbf{Imperceptibility.}
Protected inputs remain visually indistinguishable, as we adopt PAChroma's $\ell_\infty$ luminance budget and perception-aware masking strategy. Even at small perturbation budgets, the outputs remain colorful yet visibly content-inconsistent, indicating strong imperceptibility of the protected inputs to human observers (Fig.~\ref{fig:epsilon_effect}).

\textbf{Robustness.}
Figure~\ref{fig:robustness} shows that SCNB's content-inconsistent colors persist after common post-processing, whereas PAChroma's outputs often appear more plausible after these operations, exhibiting a faint but content-consistent tint.
Table~\ref{tab:imagenet_transfer_scnb_pachroma_bigcolor_source} confirms this quantitatively: CaCDD remains high even after post-processing for SCNB (e.g., for BigColor, $3.05\!\rightarrow\!2.68$ at JPEG50 and $2.89$ under RRC), indicating that the semantic color break is not a fragile artifact but persists under realistic distribution shifts.

\begin{figure}[t]
\centering
\includegraphics[width=\linewidth]{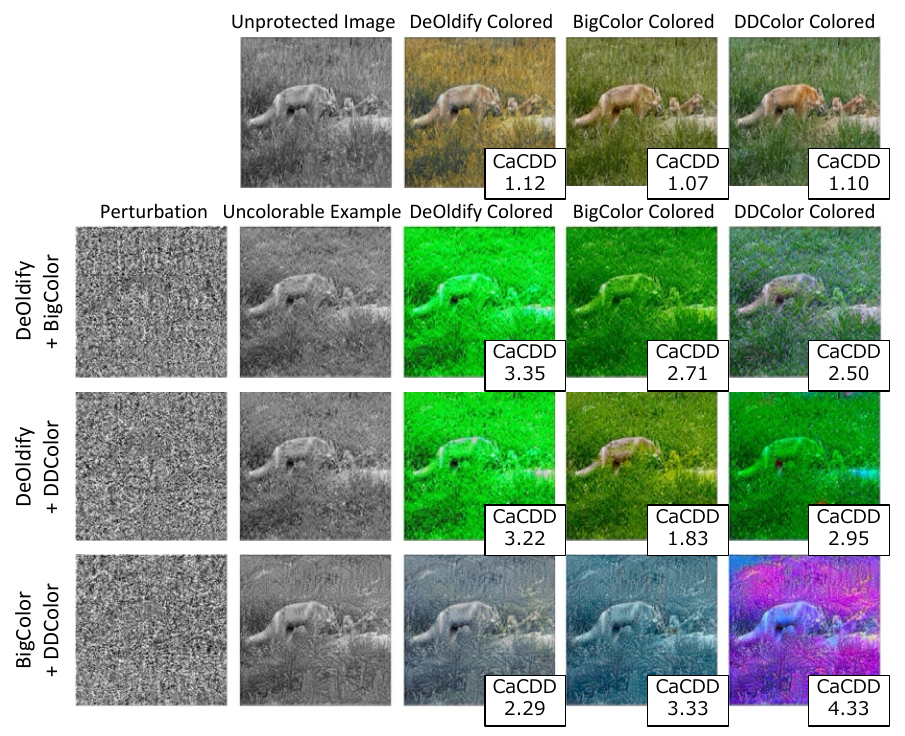}
\caption{\textbf{Improved transferability via ensemble-crafted UE.} Rows correspond to surrogate ensembles used for crafting SCNB and columns to attack (evaluation) models.}
\label{fig:ensemble}
\end{figure}

\textbf{Transferability.}
Single-source SCNB yields large CaCDD increases in the white-box setting but only modest black-box gain (Table~\ref{tab:imagenet_transfer_scnb_pachroma_bigcolor_source}), likely because perturbations optimized against one surrogate tend to exploit model-specific color decoding features that do not generalize across architectures.
Ensemble-crafted SCNB alleviates this by encouraging perturbations that exploit vulnerabilities shared across surrogates, producing more consistent CaCDD gains on unseen attack models (Fig.~\ref{fig:ensemble}, Table~\ref{tab:ensemble_cacdd_robust}).

\begin{table}[t]
\caption{\textbf{Transferability with ensemble-crafted UE.} CaCDD increases on unseen target models under $\epsilon$=32/255 when SCNB is crafted using multiple surrogate colorizers.}
\label{tab:ensemble_cacdd_robust}
\centering
\footnotesize
\renewcommand{\arraystretch}{1.08}
\setlength{\tabcolsep}{3pt}

\begin{tabular*}{\columnwidth}{@{\extracolsep{\fill}} l c c c @{}}
\toprule
Source Model & Attack model &
CaCDD$_{\text{unprot}}$ & CaCDD$_{\text{prot}}$ \\
\midrule

\multirow{3}{*}{\shortstack{DeOldify\\+ BigColor}}
& DeOldify & 1.41 &\textbf{ 2.84 (+101\%)} \\
& BigColor & 1.26 & \textbf{2.66 (+111\%)} \\
& DDColor  & 1.25 & \textbf{1.57 (+26\%)} \\
\midrule

\multirow{3}{*}{\shortstack{BigColor\\+ DDColor}}
& DeOldify & 1.41 & \textbf{1.59 (+13\%)} \\
& BigColor & 1.26 & \textbf{2.48 (+97\%)} \\
& DDColor  & 1.25 & \textbf{3.21 (+157\%)} \\
\midrule

\multirow{3}{*}{\shortstack{DeOldify\\+ DDColor}}
& DeOldify & 1.41 & \textbf{2.74 (+94\%)} \\
& BigColor & 1.26 & \textbf{1.47 (+17\%)} \\
& DDColor  & 1.25 & \textbf{3.40 (+172\%)} \\
\bottomrule
\end{tabular*}
\end{table}

\section{CONCLUSION}
We presented \textbf{SCNB}, a proactive defense that generates \emph{Uncolorable Examples} which preserve the visual appearance of published grayscale images while causing unauthorized colorizers to output \emph{content-inconsistent} colors.
Central to SCNB is \textbf{CaCDD}, a ground-truth-free, content-aware metric based on semantic color priors, used for both evaluation and optimization.
Across three representative colorizers, SCNB achieves strong semantic disruption under small perturbation budgets and remains robust to common post-processing.
Unlike PAChroma, which mainly reduces chroma magnitude and can still yield visually acceptable results, SCNB directly targets \emph{semantic} color implausibility, better discouraging illegitimate colorization of monochrome media.
Future work includes extending to videos, user-hint colorization, and evaluating stronger adaptive attackers alongside human studies.

\bibliographystyle{IEEEbib}
\bibliography{main}

\clearpage

\twocolumn[
\begin{@twocolumnfalse}

\begin{center}
{\LARGE\bfseries
Appendix for "Semantic Color Naturalness Breaker: Preventing Illegitimate Colorization via Content-Aware Color Priors"
\par}

\vspace{1em}

{\large
Yuki Nii$^{\star}$ \qquad
Futa Waseda$^{\star}$ \qquad
Ching-Chun Chang$^{\dagger}$ \qquad
Isao Echizen$^{\star\dagger}$\par}

\vspace{0.5em}

$^{\star}$The University of Tokyo, Japan \qquad
$^{\dagger}$National Institute of Informatics, Japan

\vspace{1em}
\end{center}

\vspace{1em}

\end{@twocolumnfalse}
]

This appendix complements the main paper with additional details and extended results:
(i) CaCDD prior/scoring settings used in our experiments,
(ii) CaCDD component ablations (distance vs.\ hue-direction),
(iii) analysis of CaCDD compared to conventional metrics, and
(iv) computational costs and additional results of SCNB.

\section{CaCDD implementation details}
\label{sec:supp_hparams}

\subsection{Semantic Color Priors and Mask Protocol}
\label{subsec:supp_priors}

CaCDD uses content-conditioned color priors in the ab channels of CIELAB.
For each semantic label $c\in\mathcal{C}$, we collect color samples $z=(a,b)\in\mathbb{R}^2$ from semantic regions on a large reference corpus and fit a $K$-component Gaussian mixture model (GMM):
\begin{equation}
\label{eq:prior_gmm}
p_{\text{real}}^{(c)}(z)=\sum_{k=1}^{K}\pi_{c,k}\,\mathcal{N}(z;\mu_{c,k},\Sigma_{c,k}).
\end{equation}

\textbf{Automatic semantic regions via RAM and Grounded-SAM.}
To construct label-conditioned color priors at scale without manual annotation, we use an open-vocabulary tagging--segmentation pipeline.
For each reference image, RAM (Recognize Anything Model) predicts a ranked list of semantic tags; after basic text filtering we keep the top-$m$ tags as candidate labels $c\in\mathcal{C}$ (Tab.~\ref{tab:prior_params_natural}).
For each tag, we localize the corresponding region using Grounded-SAM (text-grounded detection + segmentation), producing a pixel-accurate binary region mask $\Omega_c$.

\textbf{Mask protocol and color sampling.}
We discard low-quality masks using simple geometric and confidence-based rules (e.g., tiny area or low grounding confidence), retaining only reliable semantic regions.
For each accepted region $\Omega_c$, we convert the image to CIELAB and collect color samples $z=(a,b)$ from pixels inside the region, i.e., $\{(a_p,b_p)\mid p\in \Omega_c\}$.
To avoid a small number of images dominating the fit, we subsample at most $N_{\max}$ color points per image.
Aggregating samples over the full corpus yields a color set per label; labels with fewer than $N_{\min}$ samples are discarded.
Finally, we fit the per-label $K$-component GMM in Eq.~\eqref{eq:prior_gmm}.
Fig.~\ref{fig:supp_prior_pipeline} summarizes the construction pipeline, and
Tab.~\ref{tab:prior_params_natural} lists the hyperparameter used in our setting.

\textbf{Fallback prior and mask failures.}
If no valid semantic prior is available for a pixel (e.g., no reliable $\Omega_c$ covers it), we fall back to a global, category-agnostic color prior fit from all pooled color samples.
On ImageNet val50k, semantic mask computation succeeds for $99.3\%$ of images.

\begin{table}[t]
\centering
\caption{\textbf{Hyperparameters for constructing semantic color priors.}}
\small
\label{tab:prior_params_natural}
\renewcommand{\arraystretch}{1.15}
\setlength{\tabcolsep}{5pt}
\begin{adjustbox}{max width=\linewidth}
\begin{tabular}{@{} l l p{0.53\linewidth} c @{}}
\toprule
\textbf{Stage} & \textbf{Param.} & \textbf{Meaning} & \textbf{Value} \\
\midrule
Tagging / labels
& \texttt{$tag_{\text{topm}}$} & Max number of filtered tags used in the prompt & 25 \\
& \texttt{$tag_{\text{chunk}}$} & Chunk size for tag prompts to avoid long captions & 10 \\
\addlinespace[0.3em]

GroundingDINO
& \texttt{$box_{\text{min}}$} & Minimum box score to keep a detection & 0.30 \\
& \texttt{$text_{\text{min}}$} & Minimum text-match score to keep a detection & 0.30 \\
\addlinespace[0.3em]

\shortstack[l]{Mask quality filters\\(SAM output)}
& \texttt{$area_{\text{min}}$} & Minimum mask area ratio & 0.01 \\
& \texttt{$area_{\text{max}}$} & Maximum mask area ratio & 0.95 \\
& \texttt{$score_{\text{min}}$} & Minimum detection confidence for a mask & 0.35 \\
& \texttt{$bboxIoU_{\text{min}}$} & Minimum IoU between mask bbox and detection box & 0.30 \\
\addlinespace[0.3em]

GMM fitting
& $K$ & Number of mixture components per label & 3 \\
& $\lambda_{\text{reg}}$ & Diagonal covariance regularization ($\Sigma_{c,k}\!\leftarrow\!\Sigma_{c,k}+\lambda_{\text{reg}}I$) & $10^{-4}$ \\
& $N_{\min}$ & Minimum samples per label (below this: discard or map to global prior) & 500 \\
& $N_{\max}$ & Per-image cap on samples to avoid domination by a few images & 2{,}000 \\
\addlinespace[0.3em]

CaCDD scoring
& $\lambda_{\text{hue}}$ & Weight for hue-direction term in CaCDD & 1.0 \\
\bottomrule
\end{tabular}
\end{adjustbox}
\end{table}


\subsection{CaCDD Scoring Definition}
\label{sec:supp_cacdd_impl}

\begin{figure}[t]
\centering
\includegraphics[width=\columnwidth]{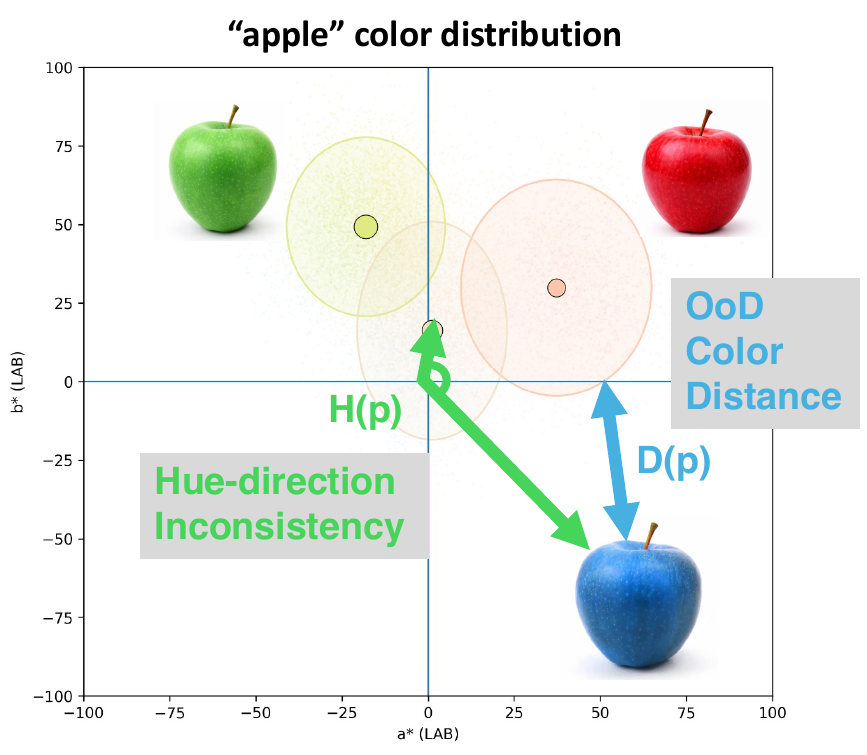}
\caption{\textbf{Illustrative idea of CaCDD components.} H(p) evaluates the inconsistency of hue directions, wheras D(p) measures the out-of-distribution distance to the GMM.}
\label{fig:cacdd_ablation_figure}
\end{figure}

CaCDD is computed on the colorization output by combining (i) Out-of-Distribution (OoD) chroma distance to the semantic GMM and
(ii) a hue-direction inconsistency term (Eq.~(2)--(7) in the main paper). Fig.~\ref{fig:supp_region_natural} reports CaCDD scores per semantic class for each image. Fig.~\ref{fig:cacdd_ablation_figure} shows an illustrative idea of the components of CaCDD. 

\section{CaCDD ablation: distance vs.\ hue-direction}
\label{sec:supp_ablation}

We evaluate the two CaCDD components by reporting:
\begin{align}
\mathrm{CaCDD}_{\text{dist}}(y) &= \frac{1}{|\Omega(y)|}\sum_{p\in\Omega(y)} D(p),\\
\mathrm{CaCDD}_{\text{hue}}(y)  &= \frac{1}{|\Omega(y)|}\sum_{p\in\Omega(y)} H(p).
\end{align}
SCNB can be optimized with either component alone by replacing the CaCDD term in the loss.

\begin{figure}[t]
\centering

\begin{subfigure}[t]{\linewidth}
  \centering
  \includegraphics[width=\linewidth,height=0.23\textheight]{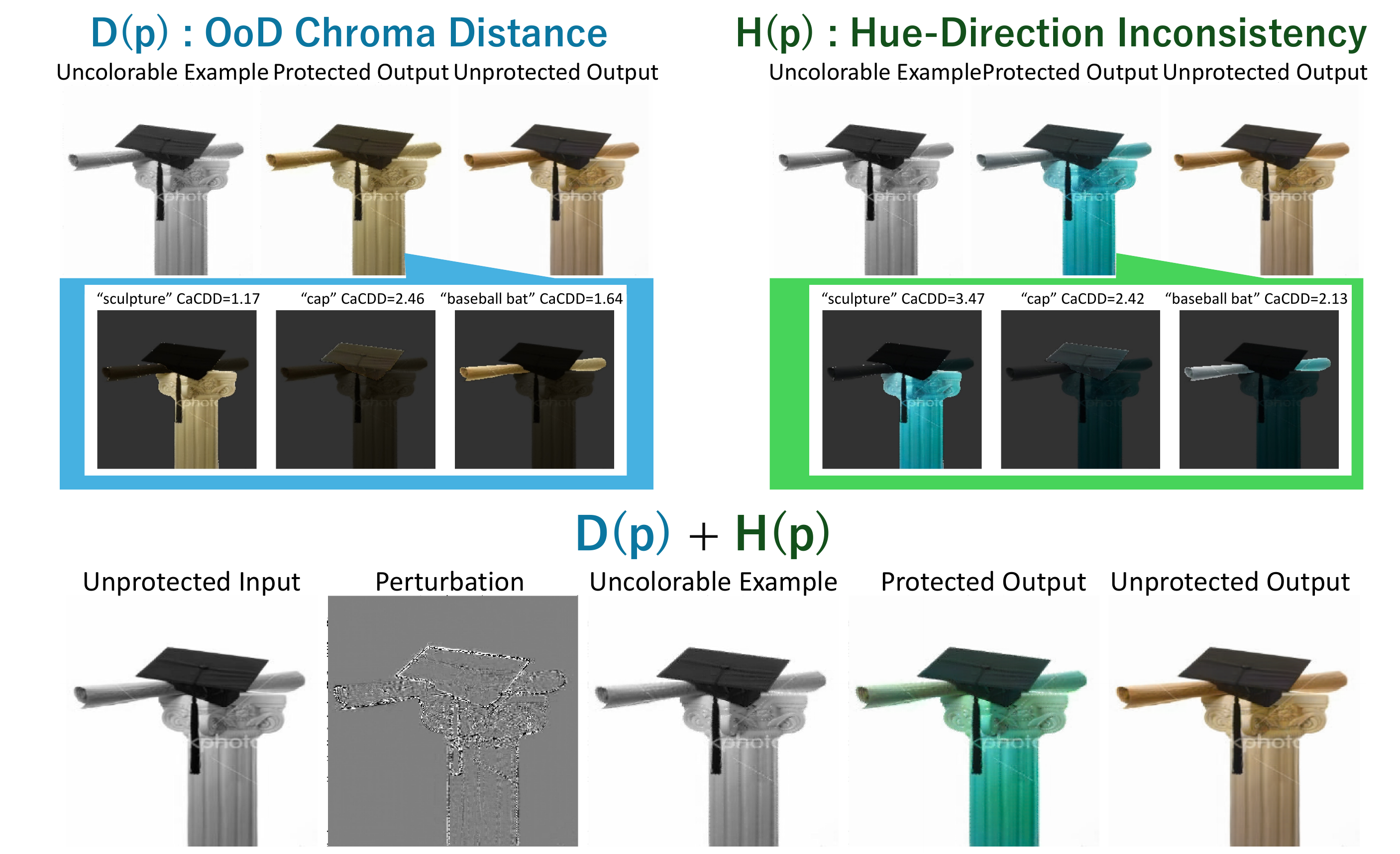}
  \caption{Optimizing only $D(p)$ yield plausible colors. Adding $H(p)$ helps steer the colors toward an inconsistent hue direction.}
  \label{fig:ablation_hue}
\end{subfigure}

\vspace{6mm}

\begin{subfigure}[t]{\linewidth}
  \centering
  \includegraphics[width=\linewidth,height=0.23\textheight]{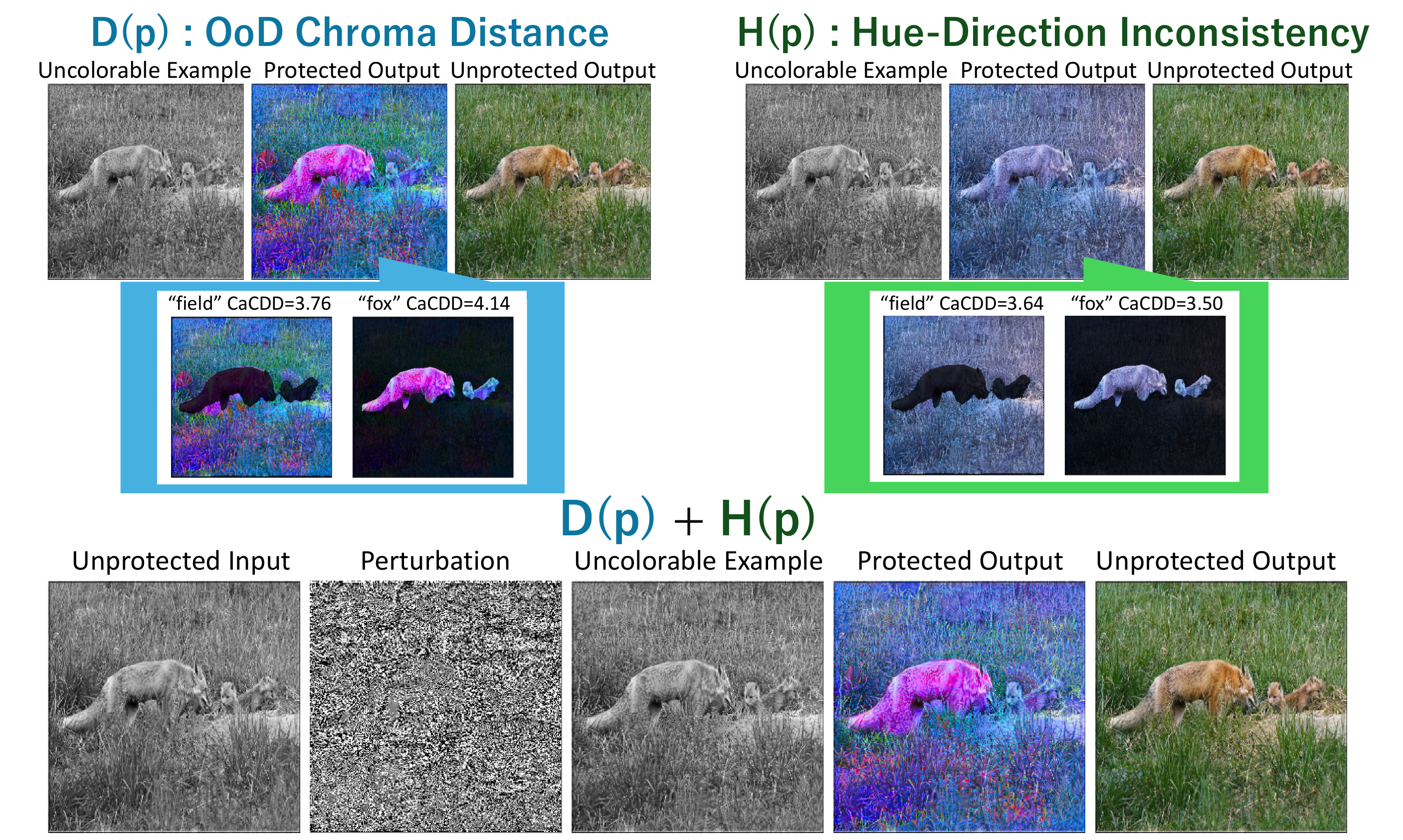}
  \caption{Optimizing only $H(p)$ yield plausible colors. Adding $D(p)$ pushes chroma away making the resulting colors more semantically implausible.}
  \label{fig:ablation_dist}
\end{subfigure}

\caption{\textbf{Ablation of CaCDD components.}
Each component alone can yield plausible-looking colors; using both terms is more consistent at producing content-inconsistent color. (Best viewed in zoom)}
\label{fig:ablation_failures}
\end{figure}

\textbf{Qualitative behavior.}
Optimizing only the distance term $\mathrm{CaCDD}_{\text{dist}}$ mainly increases out-of-distribution distance to the label-conditioned GMM (often via larger chroma magnitude), but can still produce plausible results when the change follows a visually plausible hue direction (Fig.~\ref{fig:ablation_hue}).
In contrast, optimizing only the hue-direction term $\mathrm{CaCDD}_{\text{hue}}$ tends to induce systematic hue rotations in the CIELAB $ab$ plane, but does not necessarily move chroma far from the semantic GMM in distance (Fig.~\ref{fig:ablation_dist}).
Combining both terms more reliably yields content-inconsistent color by encouraging \emph{both} (i) deviation from the semantic prior and (ii) hue-direction inconsistency.

\section{Correlation of CaCDD with existing metrics}
\label{sec:supp_corr}

\begin{figure*}[t]
\centering
\includegraphics[width=\linewidth,height=\textheight,keepaspectratio]{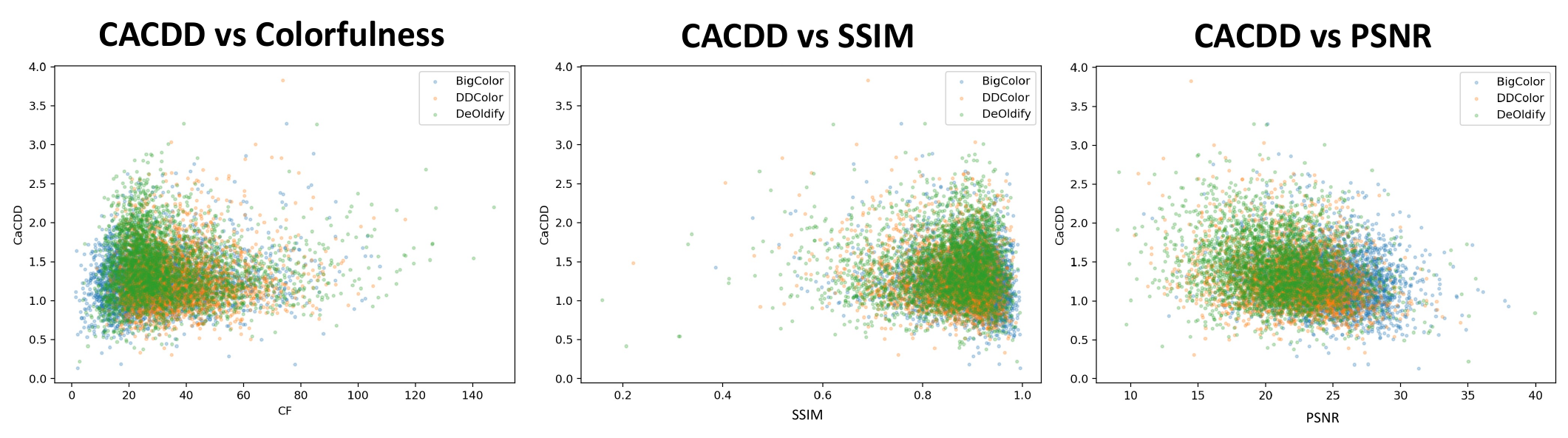}
\caption{\textbf{CaCDD show weak correlation to conventional metrics.} Scatter plots of CaCDD vs.\ CF, SSIM, and PSNR evaluated with outputs of BigColor, DDColor, and DeOldify.}
\label{fig:supp_metric_scatter}
\end{figure*}

Fig.~\ref{fig:supp_metric_scatter} and Tab~\ref{tab:imagenet_val50k_metrics} compares CaCDD against common image metrics computed on protected outputs:
colorfulness (CF), SSIM, and PSNR.
Across all three colorizers, CaCDD exhibits weak correlation with CF/PSNR/SSIM, supporting the motivation that CaCDD captures
\emph{content-conditioned color implausibility} rather than generic distortion or saturation magnitude.

\begin{table}[t]
\centering
\caption{\textbf{Evaluation of colorization models on ImageNet (val50k).}
 FID, PSNR, SSIM evaluates feature/pixel level similarity. CF measures chroma magnitude, where CaCDD demonstates semantic color plausibility.}
\label{tab:imagenet_val50k_metrics}
\renewcommand{\arraystretch}{1.15}
\setlength{\tabcolsep}{5pt}
\begin{adjustbox}{max width=\linewidth}
\begin{tabular}{@{} l c c c c c @{} }
\toprule
Method &
\multicolumn{3}{c}{\textbf{GT-required}} & 
\multicolumn{2}{c}{\textbf{GT-free}} \\
\cmidrule(lr){2-4}\cmidrule(lr){5-6}
& FID$\downarrow$ & PSNR$\uparrow$ & SSIM$\uparrow$ & CF$\uparrow$ & CaCDD$\downarrow$ \\
\midrule
DeOldify & 6.184 & 21.017 & 0.852 & 33.090 & 1.420 \\
BigColor & 2.299 & \textbf{24.775} & \textbf{0.892} & 29.188 & \textbf{1.245} \\
DDColor  & \textbf{2.233} & 22.250 & 0.865 & \textbf{35.836} & 1.269 \\
\bottomrule
\end{tabular}
\end{adjustbox}
\end{table}

\section{Additional SCNB details and results}

\subsection{Computational Cost}
\label{sec:supp_cost}
All experiments were conducted on a single NVIDIA RTX 6000 GPU.
For the reported setting, processing 40 images required approximately 13 hours in total
(about 20 minutes per image on average).

\subsection{Results Under Smaller Perturbation}
\label{sec:supp_more}
Table~\ref{tab:scnb_pachroma_cacdd} shows that SCNB remains effective even at small perturbation budgets, consistently increasing CaCDD (i.e., inducing content-inconsistent color) across attack models.
For example, on DDColor, CaCDD rises from $1.25$ to $2.30$ even at $\epsilon{=}0.0157$ (+84\%), and the elevated CaCDD largely persists after JPEG/RRC.
In contrast, PAChroma mainly suppresses chroma magnitude (low CF), which typically \emph{reduces} CaCDD toward the semantic prior (e.g., $1.41{\rightarrow}1.07$ on DeOldify at $\epsilon{=}0.1255$), indicating that desaturation alone tends to yield more semantically natural-looking outputs rather than content-inconsistent ones.

\begin{table*}[t]
\centering
\caption{\textbf{SCNB vs.\ PAChroma on ImageNet (mean over 80 runs: 40 images $\times$ 2 seeds).}
We report CF and CaCDD on unprotected/protected outputs, and CaCDD after post-processing (JPEG Q=75/50, RRC). SCNB increases CaCDD, while PAChroma reduces CF.}
\label{tab:scnb_pachroma_cacdd}

\footnotesize 
\setlength{\tabcolsep}{2.4pt}        
\renewcommand{\arraystretch}{1.02}   

\begin{tabular*}{\textwidth}{@{\extracolsep{\fill}} llc cc cc ccc}
\toprule
Method & Attack Model & $\epsilon$ &
CF$_{\text{unprot}}$ & CF$_{\text{prot}}$ &
CaCDD$_{\text{unprot}}$ & CaCDD$_{\text{prot}}$ &
CaCDD$_{\text{jpeg75}}$ & CaCDD$_{\text{jpeg50}}$ & CaCDD$_{\text{rrc}}$ \\
\midrule

\multirow{12}{*}{SCNB}
& \multirow{4}{*}{DeOldify}
& 0.1255 & 36.12 & \textbf{103.78 (+187\%)} & 1.41 & \textbf{3.00 (+113\%)} & 2.66 & 2.32 & 2.76 \\
& & 0.0627 & 36.12 & \textbf{88.50 (+145\%)}  & 1.41 & \textbf{2.73 (+94\%)}  & 2.42 & 2.14 & 2.52 \\
& & 0.0314 & 36.12 & \textbf{71.95 (+99\%)}   & 1.41 & \textbf{2.43 (+73\%)}  & 2.11 & 1.90 & 2.20 \\
& & 0.0157 & 36.12 & \textbf{54.44 (+51\%)}   & 1.41 & \textbf{2.10 (+49\%)}  & 1.82 & 1.70 & 1.86 \\
\cline{2-10}

& \multirow{4}{*}{BigColor}
& 0.1255 & 31.44 & \textbf{76.67 (+144\%)} & 1.26 & \textbf{3.05 (+142\%)} & 2.91 & 2.68 & 2.89 \\
& & 0.0627 & 31.44 & \textbf{65.82 (+109\%)} & 1.26 & \textbf{2.69 (+113\%)} & 2.49 & 2.26 & 2.57 \\
& & 0.0314 & 31.44 & \textbf{54.01 (+72\%)}  & 1.26 & \textbf{2.29 (+81\%)}  & 2.05 & 1.83 & 2.14 \\
& & 0.0157 & 31.44 & \textbf{43.11 (+37\%)}  & 1.26 & \textbf{1.89 (+50\%)}  & 1.69 & 1.53 & 1.77 \\
\cline{2-10}

& \multirow{4}{*}{DDColor}
& 0.1255 & 37.98 & \textbf{105.07 (+177\%)} & 1.25 & \textbf{3.66 (+192\%)} & 2.97 & 2.40 & 3.39 \\
& & 0.0627 & 37.98 & \textbf{91.02 (+140\%)}  & 1.25 & \textbf{3.30 (+164\%)} & 2.66 & 2.12 & 3.04 \\
& & 0.0314 & 37.98 & \textbf{72.64 (+91\%)}   & 1.25 & \textbf{2.83 (+126\%)} & 2.15 & 1.72 & 2.50 \\
& & 0.0157 & 37.98 & \textbf{54.05 (+42\%)}   & 1.25 & \textbf{2.30 (+84\%)}  & 1.68 & 1.45 & 1.97 \\

\midrule

\multirow{12}{*}{PAChroma}
& \multirow{4}{*}{DeOldify}
& 0.1255 & 36.12 & \textbf{6.33 (-82\%)}  & 1.41 & \textbf{1.07 (-24\%)} & 1.24 & 1.30 & 1.08 \\
& & 0.0627 & 36.12 & \textbf{7.23 (-80\%)}  & 1.41 & \textbf{1.10 (-22\%)} & 1.26 & 1.34 & 1.09 \\
& & 0.0314 & 36.12 & \textbf{9.68 (-73\%)}  & 1.41 & \textbf{1.16 (-18\%)} & 1.31 & 1.39 & 1.15 \\
& & 0.0157 & 36.12 & \textbf{12.98 (-64\%)} & 1.41 & \textbf{1.24 (-12\%)} & 1.37 & 1.42 & 1.22 \\
\cline{2-10}

& \multirow{4}{*}{BigColor}
& 0.1255 & 31.44 & \textbf{4.13 (-87\%)}  & 1.26 & \textbf{1.05 (-17\%)} & 1.07 & 1.06 & 1.03 \\
& & 0.0627 & 31.44 & \textbf{6.37 (-80\%)}  & 1.26 & \textbf{1.15 (-9\%)}  & 1.13 & 1.10 & 1.11 \\
& & 0.0314 & 31.44 & \textbf{10.07 (-68\%)} & 1.26 & \textbf{1.18 (-7\%)}  & 1.14 & 1.13 & 1.16 \\
& & 0.0157 & 31.44 & \textbf{16.63 (-47\%)} & 1.26 & \textbf{1.18 (-6\%)}  & 1.17 & 1.17 & 1.18 \\
\cline{2-10}

& \multirow{4}{*}{DDColor}
& 0.1255 & 37.98 & \textbf{5.00 (-87\%)}  & 1.25 & \textbf{1.07 (-15\%)} & 1.12 & 1.20 & 1.21 \\
& & 0.0627 & 37.98 & \textbf{7.79 (-79\%)}  & 1.25 & \textbf{1.18 (-6\%)}  & 1.17 & 1.18 & 1.24 \\
& & 0.0314 & 37.98 & \textbf{11.27 (-70\%)} & 1.25 & \textbf{1.21 (-3\%)}  & 1.17 & 1.18 & 1.26 \\
& & 0.0157 & 37.98 & \textbf{16.82 (-56\%)} & 1.25 & \textbf{1.28 (+2\%)}  & 1.20 & 1.20 & 1.29 \\

\bottomrule
\end{tabular*}
\end{table*}

\end{document}